\documentclass{article}
\usepackage{spconf,amsmath,graphicx,amsfonts}
\usepackage[hidelinks]{hyperref}
\usepackage{algorithm,algpseudocode}
\usepackage{booktabs}

\usepackage{subcaption}

\usepackage{color}

\title{Unconditional flow-based time series generation with equivariance-regularised latent spaces}
\name{Camilo Carvajal Reyes and   Felipe Tobar}
\address{Imperial College London}

\newtheorem{definition}{Definition}[section]

\begin{document}
\ninept
\maketitle
\begin{abstract}
Flow-based models have proven successful for time-series generation, particularly when defined in lower-dimensional latent spaces that enable efficient sampling. However, how to design latent representations with desirable equivariance properties for time-series generative modelling remains underexplored. In this work, we propose a latent flow-matching framework in which equivariance is explicitly encouraged through a simple regularisation of a pre-trained autoencoder. Specifically, we introduce an equivariance loss that enforces consistency between transformed signals and their reconstructions, and use it to fine-tune latent spaces with respect to basic time-series transformations such as translation and amplitude scaling. We show that these equivariance-regularised latent spaces improve generation quality while preserving the computational advantages of latent flow models. Experiments on multiple real-world datasets demonstrate that our approach consistently outperforms existing diffusion-based baselines in standard time-series generation metrics, while achieving orders-of-magnitude faster sampling. These results highlight the practical benefits of incorporating geometric inductive biases into latent generative models for time series.
\end{abstract}

\section{Introduction}
\label{sec:intro}

Generative modelling of time series has been extensively studied in the signal processing community. While conditioned generation can serve as a backbone for imputation and forecasting, unconditional generation is a necessary step to develop conditioned variants, as well as excelling at data augmentation. Given the success of flow-based models for continuous data, particularly diffusion models \cite{sohl-dickstein_deep_2015,song_generative_2019,ho_denoising_2020}, they have emerged as a common choice for time series generation. 

Both diffusion and flow-matching \cite{lipman_flow_2022} models work by transporting a noise source into the target (data) distribution: the former uses a stochastic differential equation, and the latter an ordinary differential equation. Despite their success in perceptual domains such as image generation \cite{dhariwal_diffusion_2021}, the fact that the sampling procedure refines the samples step-by-step means that the generation cost is often prohibitively high. Canonical alternatives to accelerate the generation are: 1) defining the models in a lower-dimensional latent space \cite{rombach_high-resolution_2022}, typically through an autoencoder (AE); and 2) adopting faster denoising paths based on ordinary differential equations (ODEs) \cite{liu_flow_2022}.  Moreover, recent works have found that equivariance of a latent space is a much desired property in the definition of reliable diffusion models for time series \cite{skorokhodov_improving_2025,kouzelis_eq-vae_2025}.

Despite the abundance of methods that make use of invariance for time series data (e.g., \cite{yang_timeclr_2022}, \cite{germain_shape_2024} and \cite{demirel_shifting_2024}), we argue that the literature is lacking a principled exploration of equivariances for flow-based time series generation. In this context, our work makes the following contributions:
\begin{enumerate}
\item We introduce a time-series generation framework that performs flow matching in a learned latent space, enabling substantially faster sampling for long sequences.
\item We demonstrate that incorporating equivariance-preserving autoencoders, trained with an explicit equivariance loss, leads to increased modelling power.
\item We show that enforcing equivariance to basic transformations, such as translation and scaling, significantly improves generation quality compared to existing flow-based baselines.

\end{enumerate}

The outline of the paper is the following: we present the basics of flow matching and autoencoders in Sec.~\ref{sec:background}; we discuss related work in Sec.~\ref{sec:related_work}; our methodology is introduced in Sec.~\ref{sec:methodology}; and the experimental evaluation is presented in Sec.~\ref{sec:experiments}.

\section{Background}
\label{sec:background}

\subsection{Flow models}
\label{sec:background:FM}

Let us consider a datapoint $x \in \mathcal{X}$, distributed according to an unknown distribution $p_{\text{data}}$. To generate samples from the unknown $p_{\text{data}}$, the (conditional) flow matching (FM) procedure \cite{lipman_flow_2022} learns a time-dependent vector field $u_t(x) : [0,1] \times \mathcal{X} \to \mathcal{X}$. This will implicitly construct a flow $\phi_t : [0,1] \times \mathcal{X} \to \mathcal{X}$ modelling the evolution of $x$ from a source distribution $p_0$ to a target distribtion $p_1$ via the following ordinary differential equation (ODE):
\begin{equation}
\label{eq:ODE_FM}
    \frac{d}{dt} \phi_t(x) = u_t(\phi_t(x)), \quad \phi_0(x) = x.
\end{equation}
In practice, the flow defines a path between a source distribution $p_0$ and a target distribution $p_1$, where it is expected that $p_1$ approximates $p_{\text{data}}$ sufficiently well. Synthesised datapoints are then generated by sampling from $p_0$, and then numerically solving the ODE to obtain samples from $p_q\approx p_{\text{data}}$. Given a distribution $q(z)$ over some conditional variable $z$ (not to be confused with a latent variable), the vector field
\[ u_t(x) := \int u_t(x|z) \frac{p_t(x|z)}{p_t(x)} q(z) \, dz \]
generates $p_t(x) = \int p_t(x|z) q(z) \, dz$ when starting from $p_0$ \cite{tong_improving_2023}. Consequently, $u_t(x|z)$ determines a probability path $\{p_t\}_{t\in[0,1]}$ that models the position of $x$ at time $t$. Given a source distribution $p_0$, from which sampling is tractable, we may choose $p_t(x|z) = p_t(x|x_0,x_1) = \mathcal{N}(x | t x_1 + (1-t) x_0, \sigma_t^2)$, so that the corresponding conditional vector field becomes $u_t(x|z) = x_1 - x_0$. A network $u_\theta(x,t)$, with parameters $\theta \in \Theta$, can then be trained to regress $u_t(x|z)$ in a supervised-learning fashion.

The choice of conditioning variable is then crucial. For instance, diffusion models correspond to the case $q(z) = p_1(x_1)$, with probability path $p_t(x|x_1) = \mathcal{N}(x|\alpha_{1-t}x_1,\sqrt{1-\alpha_{1-t}^2})$ and $p_0(x)$ fixed as a Gaussian. Previous works that consider a flexible source $q(x_0)$ mostly rely on the product $p_0(x_0)p_1(x_1)$, which assumes independence of the source and target distributions. Notice that in this case, the boundary conditions of the probability path $p_t$ are $p_1 = q_1 * \mathcal{N}(x|0, \sigma^2)$ and $p_0 = q_0 * \mathcal{N}(x|0, \sigma^2)$, where $*$ denotes the convolution operator \cite[Prop 3.3]{tong_improving_2023}. Notably, \cite{tong_improving_2023} choose $q(z)$ as the coupling $\pi(x_0, x_1)$, i.e., the optimal transport map between $p_0$ and $p_1$.

\subsection{Autoencoders}
\label{sec:background:VAE}

Autoencoders model the data probability by introducing a latent variable $z \in \mathbb{R}^l$ and modelling the joint probability $p_\theta(x, z)$. Variational autoencoders (VAE, \cite{kingma_auto-encoding_2014}) assume a standard Gaussian prior $p_0(z)$ over the latent variable, a multivariate Gaussian for $p_\theta(x|z)$ and the intractable posterior approximated by a variational posterior $q_\phi(z|x)$, with a diagonal multivariate Gaussian form. VAEs are trained by maximising the evidence lower bound (ELBO): $ \mathbb{E}_{z \sim q_\phi(z|x)}[\log p_\theta(x|z)] - D_{\text{KL}}(q_\phi(z|x) \,\|\, p(z)), $ which is a lower bound of the log likelihood $\log p_\theta(x)$. Hence, we consider a loss composed of the terms $\mathcal{L}_{\text{recon}}$ and $\mathcal{L}_{\text{KL}}$, which correspond to the negative first term and left term of the ELBO, respectively.

Given the choices of Gaussian distributions for posterior and prior, the second term has a closed form and will be referred to as the KL term. We will simplify notation by considering each autoencoder to be a pair $(\mathcal{E}_\phi, \mathcal{D}_\theta)$. The encoder $\mathcal{E}_\phi : \mathbb{R}^d \to \mathbb{R}^l$ will model the variational approximation $q_\phi(z|x)$ with neural networks that output the mean and diagonal covariance given a time series. Equivalently, $\mathcal{D}_\theta : \mathbb{R}^l \to \mathbb{R}^d$ corresponds to the decoder, modelling $p_\theta(x|z)$. The latent space dimensionality $l$ is usually chosen so that $l\ll d$, which is essential for speeding up the generation process for flow models.

Moreover, latent flow models have benefited from adding an adversarial loss to the usual VAE losses \cite{esser_taming_2021}. This means that an additional model to discriminate between real and artificial points is trained alongside the AE. The likelihood that this model assigns to reconstructions is passed as an additional (adversarial) loss, in a process inspired by generative adversarial networks. Consequently, the full VAE loss will be given by:

\begin{equation}
\label{eq:VAE_loss}
    \mathcal{L}_{\text{VAE}} = \mathcal{L}_{\text{recon}} + \beta \mathcal{L}_{\text{KL}} + \lambda \mathcal{L}_{\text{Adv}}.
\end{equation}

Here, we downweight the KL term by a factor of $0 < \beta < 1$, following related variants \cite{higgins_beta-vae_2017}. Likewise, the adversarial loss is scaled by $0 < \lambda < 1$.

\section{Related work}
\label{sec:related_work}

\subsection{Flow-based generation of time series}

Diffusion models were first used for time series generation by \cite{rasul_autoregressive_2021}, predicting future hidden states step-by-step using the diffusion formulation by \cite{ho_denoising_2020}. Other approaches, such as \cite{shen_multi-resolution_2023} and \cite{yuan_diffusion-ts_2023}, propose a diffusion-based sampling procedure that generates signals decomposed into seasonal and trend components. Moreover, \cite{tashiro_csdi_2021} and \cite{alcaraz_diffusion-based_2022} adopt diffusion for TS imputation. The use of latent diffusion has been adopted by \cite{lim_regular_2023} for unconditional generation (albeit for much shorter series), and \cite{feng_latent_2024} for forecasting. Flow matching, in particular, has also been applied in TS generation. For instance, rectified flows \cite{liu_flow_2022} are used by \cite{hu_flowts_2025} for conditional generation. On the other hand, \cite{kollovieh_flow_2025} makes use of Gaussian processes as a prior distribution, a prospect also pursued by \cite{bilos_modeling_2023} in diffusion-based generation.

\subsection{Equivariance in generative modelling}

Recent works have explored how design choices impact the convenience of a space for diffusion-based generative modelling. EQ-VAE \cite{kouzelis_eq-vae_2025} proposes an additional penalisation to autoencoder training that minimises the difference between the transformation of a reconstructed image and the reconstruction of the transformed image. This is used to force equivariance properties in the latent space. In turn, this regularisation arguably reduces the complexity of the underlying latent manifold induced by these AE. Similarly, \cite{sinha_consistency_2021} uses a regularizer that minimises the KL divergence between the encodings/reconstructions of images and transformed images, effectively promoting the invariance of the AE with respect to such operations. \cite{zhou_alias-free_2025} modify both the AE and the attention modules of latent diffusion models to improve shift-equivariance, while \cite{skorokhodov_improving_2025} argues that the modellability of the latent space is linked to its spectral properties.

\section{Equivariance regularisation for time series generative modelling}
\label{sec:methodology}

\subsection{Towards equivariant autoencoders}

We now present our proposal to induce an equivariant property in the latent space (as defined in Section \ref{sec:background:VAE}). Following the arguments presented above, this will improve the modellability of the space, hence allowing us to generate data with flow models (Section \ref{sec:background:FM}).  Given a time series $x(t) : \mathbb{R}_+ \to \mathbb{R}^d$ represented in discrete form (as evaluated at $m$ uniformly-spaced time steps), let $\mathcal{E} : \mathbb{R}^{d \times m} \to \mathbb{R}^{d_e \times l_e}$ be an encoder and $\mathcal{D} : \mathbb{R}^{l_e \times l_2} \to \mathbb{R}^{d \times m}$ a decoder.

\begin{definition}[Group equivariance]
For a transformation $g : \mathbb{R}^{d \times m} \to \mathbb{R}^{d \times m}$, we  say that $\mathcal{E}$ and $\mathcal{D}$ form a $g$-equivariant autoencoder if:
\[ \mathcal{D}(\mathcal{E}(g(x))) = g(\mathcal{D}(\mathcal{E}(x))). \]
\end{definition}

Given a group action $G$ over $\mathbb{R}^{d \times m}$
, we aim to construct a model where $\mathcal{E}$ and $\mathcal{D}$ constitute a $g$-equivariant AE for all $g \in G$. Consequently, we define the following equivariance regularisation loss:
\begin{equation}
\label{eq:eq_loss}
    \mathcal{L}_{\mathrm{eq}}(g) = \| \mathcal{D}(\mathcal{E}(g(x))) - g(\mathcal{D}(\mathcal{E}(x))) \|^2.
\end{equation}
The considered fine-tuning procedure is presented in Algorithm \ref{alg:actionEQ}.

\begin{algorithm}
\caption{Equivariance-preserving finetuning for an autoencoder }
\label{alg:actionEQ}
\begin{algorithmic}[1]
    \Require $\{x^{(i)}(t)\}_{i=1}^n$ (dataset), a pre-trained autoencoder $(\mathcal{E}, \mathcal{D})$, and a predefined action $g_p : \mathbb{R}^{d \times m} \to \mathbb{R}^{d \times m}$ of parameter $p \in [a, b]$.
    
    \For{$i = 1, \dots, n$}
        \State\textbf{Set} $p \sim \mathcal{U}([a,b])$
        \State\textbf{Set} $\mathcal{L} = \mathcal{L}_{\text{VAE}} + \mathcal{L}_{\text{eq}(g_p)},$
        \State where $\mathcal{L}_{\text{eq}}$ follows Eq.~\eqref{eq:eq_loss} and $\mathcal{L}_{\text{VAE}}$ is given by Eq.~\eqref{eq:VAE_loss}.
        \State\textbf{Do} Update autoencoder parameters of $\mathcal{E}$ and $\mathcal{D}$
    \EndFor
    \State \Return $(\mathcal{E}, \mathcal{D})$
\end{algorithmic}
\end{algorithm}

Here, $\mathcal{U}([a,b])$ refers to the uniform distribution in the interval $[a,b]$. Despite the existence of a variety of time series transformations (e.g., linear and non-linear mappings of the time variable), we show the validity of our methodology on two simple actions that act on the output of the time series:

\noindent\textbf{Translation.} Given $\delta \in \mathbb{R}$, we define the translation action as:  
\[ g_\delta(x(t)) := x(t) + \delta. \]  
This transform shifts each dimension of the series vertically. 
In general, we assume that any parameter of our action lies in an interval $[a, b)$, although we treat each dimension separately in the case of multidimensional TS.

\noindent\textbf{Amplitude Scaling.} Let $\alpha \in \mathbb{R}_+$, we consider the action $g_\alpha$ that returns the time series multiplied by $\alpha$:  
\[ g_\alpha(x(t)) := \alpha \, x(t). \] 
Similarly to the translation action, the appropriate range for the parameter $\alpha$ to be of practical use in this setting depends on the range of the signal in each dimension. We restrict $\alpha$ to lie between $0.9$ and $1.1$ for one-dimensional signals, while a larger range works for multidimensional series, e.g., $(0.3,1.7)$. This is analogous to translation, where intervals of $(-0.3,0.3)$ and $(-0.5,0.5)$ are considered, respectively.

\begin{figure}[h]
    \centering
    \begin{subfigure}{0.45\linewidth}
        \centering
        \includegraphics[width=\linewidth]{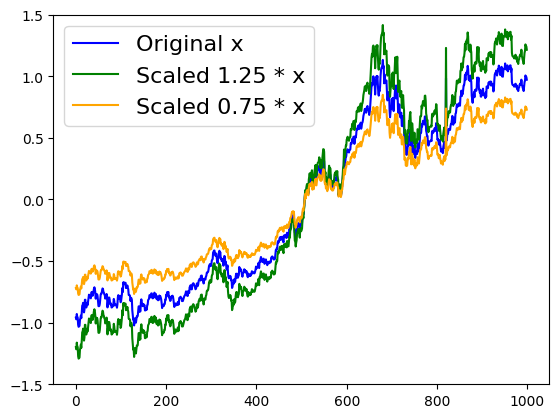}
        \caption{Scaling action.}
    \end{subfigure}
	\begin{subfigure}{0.45\linewidth}
        \centering
        \includegraphics[width=\linewidth]{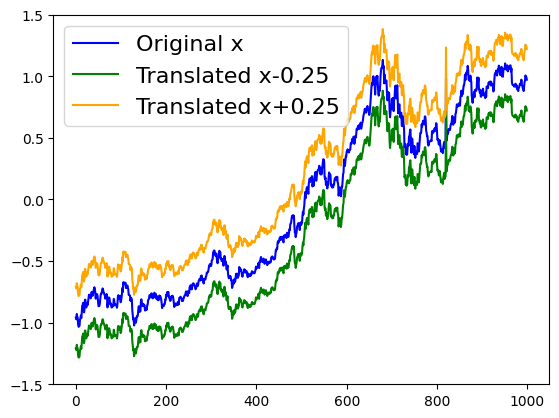}
        \caption{Translation action.}
    \end{subfigure}
    \caption{Example of our time series actions acting on a signal from the exchange rates time series \cite{lai_modeling_2018}: original first dimension in blue, along with transformed versions in green and yellow.}
    \label{fig:actions}
\end{figure}

We show an example of these time series actions in Fig.~\ref{fig:actions}.

\subsection{Generation with latent flow matching}
\label{sec:methodology:FM}
Given a latent space, latent FM trains a network $u_\theta:\mathbb{R}^l\times[0,1] \mapsto \mathbb{R}^l$, that models the dynamics between a Gaussian source and the distribution of encoded data. Following the FM setting outlined in Section \ref{sec:background:FM}, with $\sigma_t=0$, the flow will be trained to approximate the constant velocity $z_1-z_0$ from the interpolation $z_t = t z_1 + (1-t) z_0$. That is, $u_\theta(z_t,t) \approx z_1 - z_0$, where $z_0$ is the realisation of an isotropic Gaussian and $z_1$ is an encoded TS from our dataset. Consequently, we obtain a latent version of eq.~\eqref{eq:ODE_FM}:

\begin{equation}
\label{eq:latent_ODE}
    \frac{d}{dt} \phi_t(z) = u_\theta(\phi_t(z),t), \quad \phi_0(z) = z.
\end{equation}

Given a trained (latent) flow $u_\theta$, new samples are obtained by solving eq.~\eqref{eq:ODE_FM} with initial condition $z_0 \sim \mathcal{N}(0,I)$, using any standard ODE solver. Notice that FM pushes the latent samples $z_0 \sim \mathcal{N}(0,I)$ towards the distribution $q_1$ in the latent space, which satisfies $z\sim q_1$ if $\mathcal{D}(z) \sim p_{\text{data}}$.
The samples of the same VAE, but used as a generative model, correspond to decoding $z_0$. Consequently, latent FM samples are improved versions of naive VAE samples.

We claim that equivariant autoencoders are a beneficial inductive bias for generative modelling. In addition to avoiding overfitting to the training set by augmenting the dataset with plausible series, equivariant-preserving latent spaces allow for latent interpolations to translate into interpolations in the ambient space. Since FM trains the approximated vector field $u_\theta$ to regress a linear interpolation between source and target distributions, such a feature is deemed convenient.

\section{Experiments}
\label{sec:experiments}

Following the methodology described in Sec.~\ref{sec:methodology:FM}, we propose three variants: a base model, namely latent time series flow matching (LTSFM), and two fine-tuned versions that aim at preserving translation and amplitude scaling, respectively.

\subsection{Setting}

We considered three real-world datasets: 1) Household Electric Power Consumption (HEPC), containing one-dimensional voltage consumption \cite{uci_machine_learning_repository_household_2024}; 2) exchange rates, of dimension 8 \cite{lai_modeling_2018}; and 3) weather with 14 channels of weather measurements \cite{kolle_documentation_2024}. We consistently fine-tune the same base model following Algorithm \ref{alg:actionEQ} for $25$ epochs in the case of weather and HEPC datasets, and $50$ epochs for the exchange rates set. Once the FMs were trained within their respective latent spaces, generation was performed by solving the ODE in Equation \eqref{eq:ODE_FM} with either an adjoint method\footnote{\textit{torchdiffeq}: \url{https://github.com/rtqichen/torchdiffeq}} or a $10$-step fixed-step Euler method. We observe that the former yields better samples for the HEPC dataset (one-dimensional), while Euler is preferred for multidimensional series.

We tested the proposed LTSFM, before and after fine-tuning, by training convolution-based AEs for $200$ epochs (both encoder and decoder consider two convolutional layers and a linear layer). The latent dimension was set to 128 with a hidden dimension of 64. For the discriminator, we stacked three convolutional layers with a hidden dimension of 128.
The flow model follows the U-NET architecture \cite{ronneberger_u-net_2015}, where each block is a multilayer perceptron. The U-NET depth was set to two, the hidden dimensionality is $512$, and the temporal embedding (temporal in the ODE sense) corresponds to a learnable vector of dimension 128. The flows in each particular latent space were trained for $500$ epochs.

\subsection{Evaluation}

We adopted the evaluation procedure from \cite{yoon_time-series_2019} and \cite{barancikova_sigdiffusions_2024}, tailored for unconditional time series generation. The discriminative score measures the extent to which the synthetic time series can confuse a classifier, as assessed by the out-of-sample accuracy of a trained recurrent neural network (RNN). Similarly, the predictive score reports the loss of an RNN next-point predictor trained on real data, over the generated TS.
Additionally, we implemented a Kolmogorov-Smirnov (KS) test comparing the empirical distributions of test time series datasets and generated data. We report the KS score along with the percentage of points for which the null hypothesis (equal distribution) is rejected (under a $5\%$ significance). The discriminative score is reported after subtracting $0.5$ for readability, hence all metrics reflect better generation as they approach $0$.

\begin{table*}[!ht]
\caption{Discriminative, predictive and KS test metrics (KS score and reject percentage in brackets). Less is better for all metrics. The results are computed for 1000 synthetically generated time series of length 1000, compared to 1000 unseen series from the original datasets.}
\centering
\footnotesize
\label{tab:unconditional_eval}
\begin{tabular}{@{}llllllll@{}}
\toprule
dataset & Model & \begin{tabular}[c]{@{}l@{}}Discriminative\\ score\end{tabular} & \begin{tabular}[c]{@{}l@{}}Predictive\\ score\end{tabular} & \begin{tabular}[c]{@{}l@{}}KS\\ t=300\end{tabular} & \begin{tabular}[c]{@{}l@{}}KS\\ t=500\end{tabular} & \begin{tabular}[c]{@{}l@{}}KS\\ t=700\end{tabular} & \begin{tabular}[c]{@{}l@{}}KS\\ t=900\end{tabular} \\ \midrule
 & LTSFM + translation EQ (ours) & 0.086±.023 & 0.043±.003 &  0.18 (13\%) & \textbf{0.15 (5\%)} & \textbf{0.16 (4\%)} & \textbf{0.16 (6\%)} \\
 & LTSM + scaling EQ (ours) & \textbf{0.047±.028} & \textbf{0.042±.005} &  \textbf{0.17 (8\%)} & 0.17 (7\%) & 0.18 (10\%) & 0.18 (10\%) \\ \cmidrule(l){2-8} 
\multicolumn{1}{c}{HEPC} & LTSFM base model (ours) & 0.067±.035 & 0.045±.003 & 0.17 (8.4\%) & 0.24 (46\%) & 0.19 (17.6\%) & 0.25 (48.7\%) \\ \cmidrule(l){2-8} 
 & SigDiffusion & 0.070±.032 & 0.050±.012 & 0.20 (16\%) & 0.18 (9\%) & 0.19 (12\%) & 0.21 (22\%) \\
 & DDO ($\gamma$ = 1) & 0.081±.019 & 0.044±.001 & 0.25 (46\%) & 0.23 (38\%) & 0.25 (46\%) & 0.25 (51\%) \\
 & Diffusion-TS & 0.438±.057 & 0.066±.022 &  0.85 (100\%) & 0.87 (100\%) & 0.82 (100\%) & 0.88 (100\%) \\
 & CSPD-GP (RNN) & 0.415±.045 & 0.108±.002 &  0.52 (100\%) & 0.53 (100\%) & 0.55 (100\%) & 0.56 (100\%) \\
 & CSPD-GP (Transformer) & 0.500±.000 & 0.551±.028 &  1.00 (100\%) & 1.00 (100\%) & 1.00 (100\%) & 1.00 (100\%) \\
  \midrule & LTSFM + translation EQ (ours) & \textbf{0.034±.015} & 0.027±.002 & \textbf{0.16 (6\%)} & \textbf{0.16 (9\%)} & \textbf{0.16 (7\%)} & \textbf{0.16 (7\%)} \\
 & LTSM + scaling EQ (ours) & 0.046±.025 & 0.029±.002 & 0.16 (8\%) & 0.17 (11\%) & 0.17 (11\%) & 0.17 (10\%) \\ \cmidrule(l){2-8} 
\multicolumn{1}{c}{\begin{tabular}[c]{@{}c@{}}Exchange rates\end{tabular}} & LTSFM base model (ours) & 0.060±.019 & \textbf{0.026±.002} & 0.17 (8\%) & 0.18 (13\%) & 0.17 (9\%) & 0.17 (12\%) \\ \cmidrule(l){2-8} 
 & SigDiffusion & 0.278±.062 & 0.057±.001 & 0.31 (80\%) & 0.28 (67\%) & 0.28 (65\%) & 0.31 (74\%) \\
 & DDO ($\gamma=1$) & 0.326±.102 & 0.094±.004 & 0.24 (41\%) & 0.24 (42\%) & 0.25 (45\%) & 0.25 (45\%) \\
 & Diffusion-TS & 0.401±.196 & 0.120±.016 & 0.72 (100\%) & 0.71 (100\%) & 0.70 (100\%) & 0.69 (100\%) \\
 & CSPD-GP (RNN) & 0.500±.001 & 0.273±.100 & 0.59 (100\%) & 0.56 (100\%) & 0.55 (100\%) & 0.56 (100\%) \\
 & CSPD-GP (Transformer) & 0.500±.000 & 0.432±.074 & 1.00 (100\%) & 0.99 (100\%) & 0.98 (100\%) & 0.99 (100\%) \\ \midrule
 & LTSFM + translation EQ (ours) & 0.289±.112 & 0.161±.002 & 0.21 (18\%) & 0.22 (19\%) & 0.22 (22\%) & 0.23 (22\%) \\
 & LTSM + scaling EQ (ours) & 0.259±.009 & 0.158±.001 & \textbf{0.20 (15\%)} & \textbf{0.21 (19\%)} & \textbf{0.21 (19\%)} & \textbf{0.21 (22\%)} \\ \cmidrule(l){2-8} 
\multicolumn{1}{c}{Weather} & LTSFM base model (ours) & \textbf{0.248±.185} & \textbf{0.157±.002} & 0.24 (24\%) & 0.24 (26\%) & 0.24 (24\%) & 0.24 (25\%) \\ \cmidrule(l){2-8} 
 & SigDiffusion & 0.350±.080 & 0.168±.001 & 0.35 (82\%) & 0.34 (80\%) & 0.33 (76\%) & 0.34 (78\%) \\
 & DDO ($\gamma$ = 1) & 0.356±.196 & 0.307±.007 & 0.26 (45\%) & 0.27 (52\%) & 0.26 (46\%) & 0.26 (47\%) \\
 & Diffusion-TS & 0.498±.003 & 0.438±.035 & 0.49 (100\%) & 0.50 (100\%) & 0.50 (100\%) & 0.49 (100\%) \\
 & CSPD-GP (RNN) & 0.500±.000 & 0.505±.007 & 0.57 (100\%) & 0.56 (100\%) & 0.56 (100\%) & 0.56 (100\%) \\
 & CSPD-GP (Transformer) & 0.500±.000 & 0.490±.000 & 0.91 (100\%) & 0.92 (100\%) & 0.91 (100\%) & 0.91 (100\%) \\ \bottomrule
\end{tabular}
\end{table*}

\begin{figure*}[!ht]
\centering
\begin{subfigure}{0.28\textwidth}
    \centering
    \includegraphics[width=\linewidth]{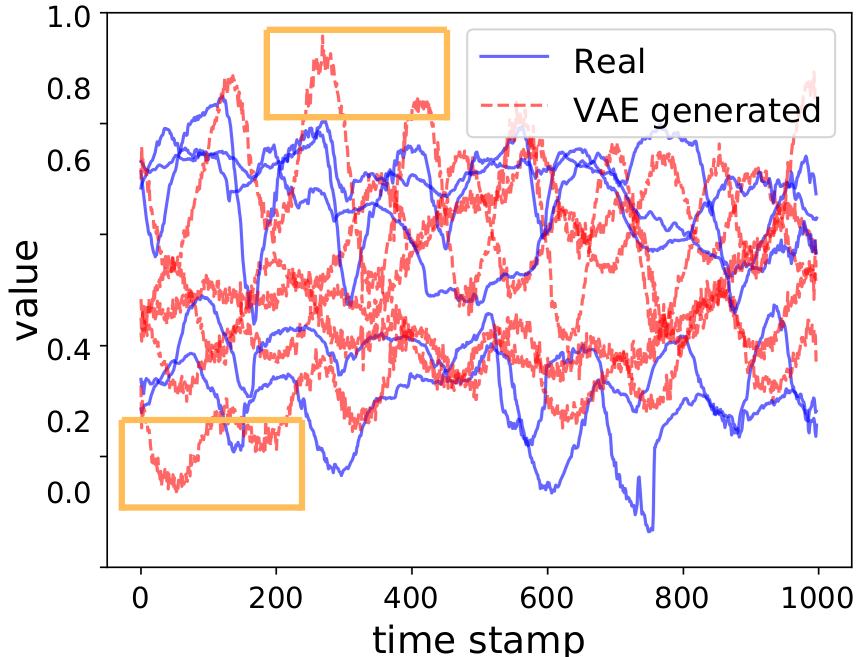}
    \caption{VAE}
\end{subfigure}
\begin{subfigure}{0.28\textwidth}
    \centering
    \includegraphics[width=\linewidth]{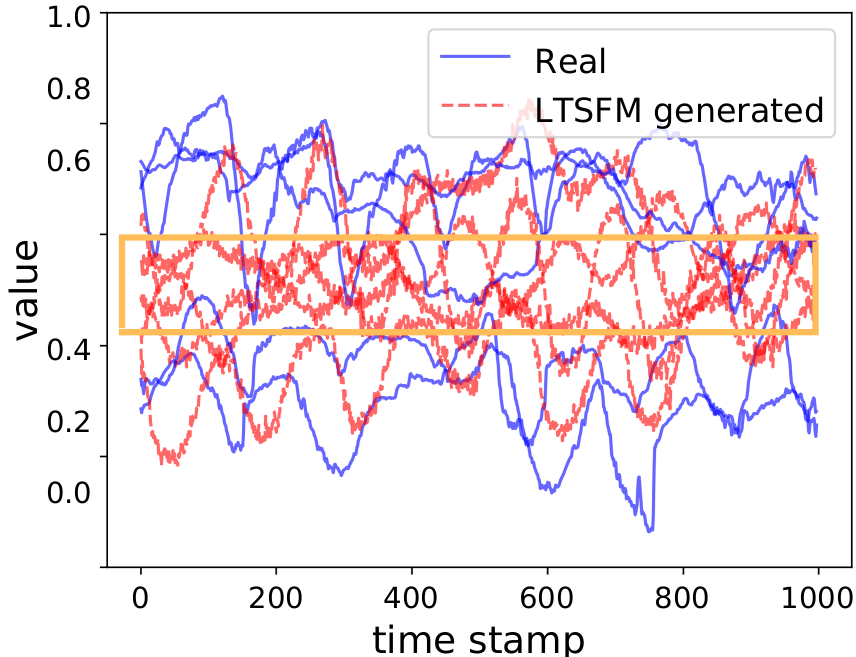}
    \caption{LTSFM base model}
\end{subfigure}
\begin{subfigure}{0.28\textwidth}
    \centering
    \includegraphics[width=\linewidth]{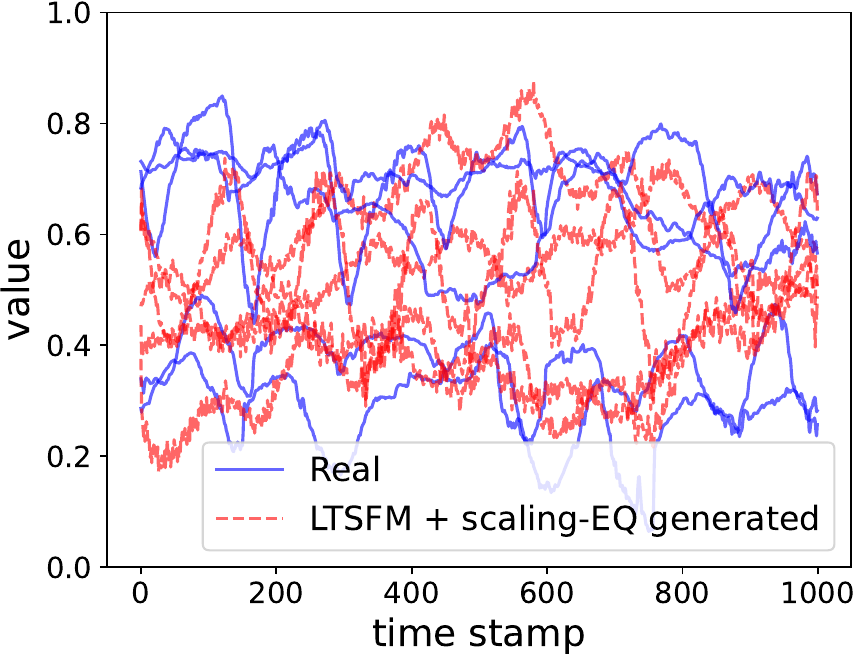}
    \caption{LTSFM + scaling regularisation}
\end{subfigure}
\caption{Synthetic samples for weather dataset \cite{kolle_documentation_2024} (dimension =10) using VAE and the proposed LTSFM and scaling-LTSFM models. The orange boxes indicate relevant undesired artefacts in the synthetic signals. }
\label{fig:flow_weather}
\end{figure*}

Table \ref{tab:unconditional_eval} shows all the performance figures considered for LTSFM alongside four diffusion-based approaches for comparison: SigDiffusion \cite{barancikova_sigdiffusions_2024}, diffusion-TS \cite{yuan_diffusion-ts_2023}, CSPD-GP \cite{bilos_modeling_2023}, DDO \cite{lim_regular_2023}. Notice the significant improvement of LTSFM and its variants over the considered state-of-the-art diffusion approaches. Critically, our model achieved better discriminative and predictive scores, while the equivariance-regularised variants improved over the base model, particularly for the KS score. Examples of generated and real time series are shown in Fig.~\ref{fig:flow_weather}, which shows how artefacts of both naive-VAE samples (large variance, often falling out of the true domain) and the base model (values too close to the mean) are less present in regularised samples, despite a slight increase in sample noise.

Furthermore, while our models take approximately 8 mins to train (2.5 mins for the AE training, 0.5 mins for fine-tuning and between 2 and 5 mins for the flow model) on an RTX3090 GPU, the generation of 1000 series of length 1000 takes less than one second. This shows an outstanding improvement over existing models, which take considerably longer for the same generation (around 11 secs for SigDiffusions and $1-42$ minutes for the other baselines), while having comparable or sometimes longer training times.
\section{Conclusions}
\label{sec:conclusions}
We have introduced the  use of autoencoders to accelerate unconditional flow-based signal generation, and shown that this methodology allows for incorporating equivariance robustness that improves out-of-distribution metrics while retaining computational efficiency.

While regularisation strategies as the proposed one do push the model towards a lower equivariance errors, strict group equivariance is not guaranteed. Therefore, building representations that allow for provable equivariance may present a useful prospect. Previous works achieving invariance of TS group actions, such as \cite{demirel_shifting_2024} or \cite{germain_shape_2024} should serve as a starting point for thse extensions. Lastly, our contribution confirms that equivariance is a direction worth exploring in generative modelling of time series.

\bibliographystyle{IEEEbib}
\bibliography{references_abbrv.bib}

\end{document}